\documentclass[pdflatex,sn-nature]{sn-jnl}% Style for submissions to Nature Portfolio journals
\usepackage{graphicx}%
\usepackage{multirow}%
\usepackage{amsmath,amssymb,amsfonts}%
\usepackage{amsthm}%
\usepackage{mathrsfs}%
\usepackage[title]{appendix}%
\usepackage{xcolor}%
\usepackage{textcomp}%
\usepackage{manyfoot}%
\usepackage{booktabs}%
\usepackage{algorithm}%
\usepackage{algorithmicx}%
\usepackage{algpseudocode}%
\usepackage{listings}%
\usepackage{subcaption}
\usepackage{tikz}
\usetikzlibrary{shapes,arrows,arrows.meta,positioning,fit}
\usepackage[final]{changes} 

\theoremstyle{thmstyleone}%
\theoremstyle{thmstyletwo}%

\theoremstyle{thmstylethree}%

\begin{document}

%\title[Human Role in LLM-Assisted Test Scoring]{Finding the Most Valuable Human Role in LLM-Assisted Written Test Scoring.}

%\title[Human Role in LLM-Assisted Test Scoring]{A Human-in-the-Loop Framework for AI-Assisted Scoring in Large-Scale Writing Assessment}
\title{A Human-in-the-Loop Framework for AI-Assisted Scoring in Large-Scale Writing Assessment}
%A Human-in-the-Loop Framework for AI-Assisted Scoring in Standardized Writing Assessment

%%=============================================================%%
%% GivenName	-> \fnm{Joergen W.}
%% Particle	-> \spfx{van der} -> surname prefix
%% FamilyName	-> \sur{Ploeg}
%% Suffix	-> \sfx{IV}
%% \author*[1,2]{\fnm{Joergen W.} \spfx{van der} \sur{Ploeg} 
%%  \sfx{IV}}\email{iauthor@gmail.com}
%%=============================================================%%

%\author{\fnm{Anonymous} \sur{Authors}}

\author[1]{\fnm{María Eugenia} \sur{Curi}}
\author*[1]{\fnm{Germán} \sur{Capdehourat}}\email{gcapdehourat@ceibal.edu.uy}
\author[1]{\fnm{Isabel} \sur{Amigo}}

\author[2]{\fnm{Magdalena} \sur{Romano}}
\author[2]{\fnm{Rosana} \sur{Serra}}
\author[2]{\fnm{Adrián} \sur{Silveira}}
\author[2]{\fnm{Andrés} \sur{Peri}}

%\author[1,2]{\fnm{Third} \sur{Author}}\email{iiiauthor@gmail.com}
%\equalcont{These authors contributed equally to this work.}

%\Author{María Eugenia Curi $^{1}$, Germán Capdehourat $^{1,}$*, Isabel Amigo $^{1}$, Magdalena Romano $^{2}$, Rosana Serra $^{2}$, Adrián Silveira $^{2}$ and Andrés Peri $^{2}$}

\affil[1]{\orgname{Ceibal}, \orgaddress{\street{Avda. Italia 6201}, \city{Montevideo}, \postcode{11500}, \country{Uruguay}}}

\affil[2]{\orgname{ANEP}, \orgaddress{\street{Av. Libertador 1409}, \city{Montevideo}, \postcode{11100}, \country{Uruguay}}}

\abstract{The integration of artificial intelligence (AI), particularly large language models (LLMs), into educational assessment has opened new opportunities to enhance the efficiency and scalability of grading processes. This study presents the design and validation of an AI-assisted scoring framework for written responses in a large-scale national assessment. The proposed approach focuses on short written texts of approximately 150-200 words and incorporates a human-in-the-loop strategy to preserve assessment quality while reducing manual workload.
The study is grounded in a real operational context, using data from two recent editions of a nationwide test, each comprising approximately 5,000 student responses. We analyze the alignment between AI-generated scores and human raters across multiple rubric dimensions, as well as the impact of the proposed decision flow on pass/fail outcomes. Results show moderate to high agreement between the model and human evaluations in most dimensions, supporting the feasibility of AI assistance in this setting. Moreover, the proposed correction workflow identifies cases where human review is most valuable, enabling a more efficient allocation of expert effort.
The findings suggest that AI-assisted scoring can be safely integrated into large-scale assessment processes \added{only} when combined with carefully designed human oversight. The paper concludes by discussing practical implications for deployment in national assessment systems and outlining future research directions, including longitudinal monitoring of model-human alignment and the analysis of potential cognitive bias introduced by AI-supported review workflows.}


\maketitle

\section{Introduction}\label{sec:intro}
The recent advance of large language models (LLMs) has renewed interest in the use of artificial intelligence (AI) to support educational assessment processes, particularly in tasks involving written production \cite{jukiewicz_can_2026}. Automated essay scoring has been studied for decades; however, recent advances in generative AI have significantly expanded the range of feasible approaches, especially for short argumentative texts that require holistic and rubric-based evaluation \cite{mizumoto_exploring_2023}. Despite these advances, evidence from real large-scale operational contexts remains limited, and important questions persist regarding reliability, validity, and the appropriate role of human oversight \cite{jukiewicz_can_2026}.

This study presents the integration of LLM-based scoring into the evaluation process of Spanish written productions within a national large-scale assessment. The target task involves short argumentative texts of approximately 150-200 words, evaluated using a detailed analytic rubric by trained human raters. The assessment constitutes a high-stakes certification pathway for over-age students seeking to complete lower secondary education, which places strong requirements on scoring quality, consistency, and fairness.

A key strength of this work lies in the availability of operational data at scale. The study draws on two recent editions of the national test (2024 and 2025), each comprising approximately 5,000--6,000 participants. This setting provides a rare opportunity to examine AI-assisted scoring under realistic conditions, including established human scoring workflows, expert-designed rubrics, and longitudinal comparability across test forms.

The objective of this research is to design and validate an AI-assisted assessment process that can be integrated into the existing evaluation workflow. Beyond measuring raw agreement between AI and human raters, the study investigates how AI-based scoring behaves within the full decision pipeline, including proficiency-level classification and pass--fail outcomes. In addition, the work explores the consistency of the model, its generalization across test editions, and its potential to reduce human scoring workload while preserving decision quality.

Based on the empirical findings, we propose a Human-in-the-Loop (HITL) evaluation framework that strategically combines automated scoring with expert human review. The proposed approach aims to accelerate result reporting, optimize the use of human expertise, and maintain the quality standards required in high-stakes assessment contexts.
\added{
To guide the empirical evaluation, the study addresses the following research questions:}

\textbf{\added{RQ1:}} \added{To what extent can a large language model, guided by detailed prompting, achieve agreement with expert human raters in the assessment of argumentative writing in Spanish within a large-scale national examination?}

\textbf{\added{RQ2:}} \added{Can a Human-in-the-Loop (HITL) evaluation framework be designed so that potentially critical AI scoring errors are systematically identified and reviewed by expert human raters, thereby preserving the fairness and reliability required for high-stakes pass/fail decisions?}

\textbf{\added{RQ3:}} \added{To what extent can AI-assisted scoring reduce human scoring workload while maintaining the quality of assessment decisions?}

The main contributions of this paper are the following:

\begin{itemize}
    \item An empirical evaluation of LLM-based scoring in a real national large-scale writing assessment.
    \item A cross-year analysis (2024--2025) examining robustness and generalization of prompt-based scoring.
    \item A detailed comparison between AI scoring and human inter-rater agreement at the rubric-item level.
    \item The design and simulation of a Human-in-the-Loop operational framework for AI-assisted scoring.
    \item An estimation of the potential reduction in human scoring workload under realistic deployment assumptions.
\end{itemize}

Overall, this work contributes new evidence on the practical integration of LLMs into high-stakes educational assessment and outlines a feasible pathway toward responsible, scalable AI-assisted evaluation.

%\section{Related Work}\label{sec:rel_work}
%\input{rel_work.tex}

\section{\replaced{Literature Review and Research Positioning}{Related Work}}
\label{sec:rel_work}
Automatic evaluation of student responses has been the subject of research for many years. Several solutions based on NLP techniques, prior GPT technologies, were explored and proposed \cite{burrows_eras_2015}. In recent years, the subject has regained attention, since LLMs provide not only the possibility to increase accuracy in those cases where previous results exist, but also promise to expand the domains of application, as well as allowing for personalized and timely feedback. This ultimately results in a significant reduction in teacher workload, while fostering the improvement of learning outcomes. All of this without requiring labeled data for training. 

Automated grading has been proposed for various educational settings. Indeed, a large set of use cases focus on the linguistic domain. Evaluations are in the vast majority text-based, and might include short answers, open-ended questions and essays. Most studies focus on English language, see for instance \cite{li_evaluating_2024} for a recent result. Further text-based use-cases involve computer science domain \cite{pecuchova_automated_2025} and science \cite{deng_rubric-conditioned_2026}, with databases that are usually in English. Fewer studies focus on the chemistry (see e.g. \cite{cvengros_assisting_2025}) and mathematics domains (see e.g. \cite{caraeni_evaluating_2024}), where  handwritten responses are usually given, and evaluation involves combining visual and textual information, thus being more complex to handle. A whole other range of use cases involve video or audio automatic evaluation, we shall leave out of the scope of this review such cases. We also focus on recent studies, as the rapid advances of technology capture the more relevant solutions.  %%AcÁ citar todos los diferentes dominios y sus papers 

\added{Research specifically addressing AI-assisted grading of Spanish writing remains limited. Existing studies have primarily focused on proof-of-concept evaluations using relatively small datasets or educational settings with limited operational constraints (e.g.~\cite{capdehourat2025}), and few have investigated rubric-based assessment of argumentative writing in large-scale examinations (e.g.~\cite{Zimotti2026}). As a result, evidence regarding the applicability of LLM-assisted grading to high-stakes Spanish-language assessments is still scarce, leaving important questions about reliability, operational deployment, and human oversight largely unexplored.}

In this context, we are particularly interested in reviewing related work to gain insights on to what extent have previous results succeeded in these automated tasks, what metrics have they used in order to establish validation of the method and which techniques have been used to optimize such metrics. In addition, how human oversight is addressed by the proposed methods is of paramount importance to our real use case, as well as what human perceptions about introducing AI the evaluation flow are. It is with these different lenses that we analyze and summarize related work in the following lines. 

%\subsection{The techniques used for LLM assisted grading}
\subsection{Techniques for LLM-assisted grading}

In \cite{jukiewicz_can_2026}, an exhaustive and recent review on automated grading is presented. The authors reviewed 42 papers, out of which the most frequently represented disciplines are computer science and foreign languages at the top, followed by mathematics and medicine and fewer cases of engineering and finally social sciences.

This plethora of use cases makes the related literature rich, although the results are not easy to generalize. \replaced{Studies on Spanish writing assessment remain comparatively scarce and are mostly limited to specific educational settings, such as Spanish as a second language~\cite{Zimotti2026}. To the best of our knowledge, no previous work has investigated LLM-assisted grading of Spanish argumentative writing within a large-scale, high-stakes assessment.}{Moreover, to the best of our knowledge, no work has focused on our case study: short argumentative Spanish texts.}
Not surprisingly, \cite{jukiewicz_can_2026} highlights that LLMs work best in English, although they are evolving and increasing their capabilities in other languages. Also, according to the same review, results are better when rubrics are used, or examples are given, and when tasks require short and structured answers, rather than long ones or where opinion is involved. For use cases more similar to ours, \cite{jukiewicz_can_2026} concludes that expert oversight is necessary, along with detailed rubrics and carefully defined prompts. This is also the approach of the framework proposed in \cite{Selvam_2025}. \cite{caraeni_evaluating_2024}, \cite{chu_llm-based_2025} and \cite{sonkar_automated_2024} are also examples where the need of well designed and detailed rubric is emphasized. % In \cite{steks}, authors show that breaking down grading into smaller rubric items helps models focus on smaller parts of the task instead of doing the entire task altogether. 
 The need of human oversight is indeed pointed out by several studies, not only for legal and ethical reasons but also for performance ones (see e.g. \cite{cvengros_assisting_2025} and \cite{jukiewicz_can_2026}). How to measure performance and when to trigger human intervention is addressed in the following. 

 %The review concludes that automated LLMs can accelerate the grading process, while the best results are obtained when they are used along with human oversight. \cite{jukiewicz_can_2026} also highlights that LLMs work best in English, although they are evolving towards increasing their capabilities in other languages. Also, according to the same review, results are better when rubrics are used, or examples are given, and when tasks require short and structured answers, rather than long ones or where opinion is involved.  In \cite{steks}, authors show that breaking down grading into smaller rubric items helps models focus on smaller parts of the task instead of doing the entire task altogether. 
  
%% metricas y succes rates
%\subsection{The metrics used to validate LLM assisted grading solutions}
\subsection{Metrics for validating LLM-assisted grading systems}

Accuracy is one of the most used metrics in the literature and is defined as the  level of agreement between AI and human expert evaluators, with some variations in order to capture non random agreements such as Cohen’s Kappa and Quadratic Weighted Kappa (QWK) to adjust for chance agreement, as well as Fleiss’ Kappa and Krippendorff’s Alpha for multi-rater scenarios. In addition to accuracy or precision, some articles use standard metrics such as F1 score, recall, or even normed F1, to handle highly granulated grades  \cite{cvengros_assisting_2025}. \cite{craig_ai-marking_2025} highlights that beyond accuracy, for ethical and legal reasons, it is of paramount importance to take into account metrics based on false positives or false negatives, which translates into measuring the performance of the method in terms of over-grading and under-grading. Indeed, under-grading poses equity concerns as it unfairly penalizes students, for instance with conceptually correct answers but with atypical phrasing \cite{pecuchova_automated_2025}, while over-grading are errors that are difficult to detect, as it is less likely that an audit is requested by students in this case \cite{cvengros_assisting_2025}.  %In our setting, we shall capture these aspects by taking into account the details of the whole test and analyzing final decisions (pass vs fail).

%% mencionar certidumbre y nuestra consistencia
Other metrics aim to capture trust in the evaluation model, or in other words, the reliability of the prediction. This concept is frequently referred as \textit{uncertainty}. In \cite{li_how_2026} a comprehensive benchmark of different uncertainty metrics is performed for different settings. The study provides valuable insights and concludes that no metric is universally optimal, and selecting the proper one should be carefully studied in each setting. Among the insights, they highlight that categoric metrics have shown to be more effective in capturing uncertainty than semantic ones. In the following, we illustrate different definitions of uncertainty metrics found in the literature. %We focus on Semantic Entropy, used in \cite{iyer_towards_2025},  Max-Agree-Rate (MAR) \cite{li_how_2026}, the Area Under the Accuracy-Rejection Curve, and method based on Item Response Theory (IRT) \cite{cvengros_assisting_2025}. Finally, a more practical approach is used in \cite{yang_pensieve_2025}, where the result of three models is compared, and agreement is assumed to imply trust.
%is used to define a Risk metric, which quantifies the deviation between AI scores and psychometric expectations to determine the need for human intervention. 

Typically, uncertainty metrics are obtained by calculating some index based on the output of querying the model repeatedly with the same input. 
Semantic metrics group the output according to their semantic meaning. In particular, semantic entropy (see e.g. \cite{iyer_towards_2025}) is based on querying the model several times to generate explanations to the same response. Then explanations are clustered according to their meaning. Finally, the probability of each cluster is calculated, and the Shannon formula for entropy is applied. High values of entropy are interpreted as a signal of either poor rubric specification, difficult to understand response, or high uncertainty of the model. Low values of entropy, in turn, are obtained when all explanations are fairly similar semantically, and interpreted as a signal of confidence on the model output. The method is elegant and as shown in \cite{iyer_towards_2025} it can capture uncertainty quite well, though it also fails when the LLM is confident but incorrect. Also, results do not necessarily generalize to different settings. On the other hand, categoric approaches rely on the labels or grades given by the model, rather than semantic explanations. Categoric entropy \cite{li_how_2026} is calculated in a way similar to Semantic entropy but frequency of grades are considered (rather than concepts). A simpler version is Max-Agree-Rate (MAR) \cite{li_how_2026}, that measures the proportion of the most frequent grade. A large MAR value is obtained when the model is consistent and is interpreted as a signal of low uncertainty. A low value signals high uncertainty.

Yet another approach is the one based on Item Response Theory (IRT). Rather than measuring uncertainty related to the model's internal consistency, this approach measures how ``surprising" is one output based on the student demonstrated abilities and the item difficulty \cite{cvengros_assisting_2025}. The uncertainty is then measured as the difference of the expected grade (obtained from the IRT modelling) and the grade given by the LLM.

Regardless of the way of calculating uncertainty, its effectiveness must be carefully study for each particular setting.  Effectiveness of an uncertainty metric can be measured using for instance methods like \replaced{AUC or C-Index}{AUROC, AUARC or C-INDEX}. \replaced{AUC}{AUROC} (Area Under the Receiver Operating Characteristic curve) \cite{xia_survey_2025} measures the probability that a randomly chosen incorrect response is assigned an uncertainty value higher than a randomly chosen correct response. The higher the \replaced{AUC}{AUROC} value, the better the metric is at discriminating reliable from unreliable model outputs. The C-Index (Concordance Index introduced in \cite{steck_ranking_2007}) is a rank-based metric, it assesses whether responses that have larger true errors also receive higher uncertainty scores. The higher the C-Index the more effective the uncertainty metric is.

%%CREO que está bueno decir que no usamos estos mopdelos porque no se sabe como generalizan, y ademas lo que hacemos es no dejar que nadie pierda sin mirarlo. y es caro estar haciendo consistency todo el tiempo. igual podríamos ver si no tenemos todo para aplicar IRT. 

Beyond accuracy and uncertainty, other metrics are found in the literature that are worth citing. For instance, robustness, defined as sensitivity to synonyms in prompts and to prompt injection, is measured in  \cite{deng_rubric-conditioned_2026}. Beyond technical metrics, some works focus on measuring explainability and transparency of outputs of the model, for instance using SHAP (see \cite{pasupuleti_human---loop_2025} and \cite{Selvam_2025}), these studies also highlight that human in the loop approaches increase transparency. Finally, the time saved by automated grading  is measured in \cite{craig_ai-marking_2025}. 

\section{\replaced{Research Gap and Contribution}{Positioning of the Present Study}}

Taken together, the literature provides a rich and nuanced set of metrics for evaluating LLM-assisted grading systems, ranging from agreement-based measures to more sophisticated approaches for estimating uncertainty, robustness, and explainability. 
\replaced{However, comparatively little attention has been devoted to how these models can be safely integrated into operational assessment workflows, particularly in high-stakes contexts where grading outcomes directly determine certification decisions. Likewise, evidence on large-scale Spanish-language assessments remains scarce. Consequently, an important research gap lies not only in achieving high agreement with expert raters, but also in designing deployment strategies that preserve fairness and reliability despite the inevitable errors of AI models.
}{However, these metrics are typically designed to assess model performance in isolation, focusing on the intrinsic reliability of predictions rather than on their impact within a broader assessment pipeline.}
In large-scale, high-stakes evaluation settings such as the one considered in this study, the practical relevance of a given metric depends not only on its theoretical properties but also on how it informs operational decisions within the overall assessment process.

In this work, while we draw on standard metrics such as accuracy, inter-rater agreement, and consistency to validate the alignment between AI and human raters, our primary focus lies elsewhere. Specifically, the evaluation perspective is shifted from model-centric performance to decision-oriented impact, analyzing how AI-generated scores affect final outcomes when integrated with other components of a standardized test, such as multiple-choice sections. This approach prioritizes the identification of cases in which AI predictions may alter pass/fail decisions, thereby requiring human review, rather than relying solely on uncertainty estimates to determine model trustworthiness. In doing so, the work emphasizes the design of an operational workflow that balances efficiency and reliability in real-world conditions, aligning evaluation practices with the ultimate goal of fair and accurate decision-making.

All in all, automated evaluation based on LLMs and well-designed rubrics and prompts has shown promising results for English written assessments. However, as aforementioned, relatively few studies focus on Spanish texts. Given that LLMs are not equally trained across different languages, findings from English-based studies are not necessarily transferable to other contexts, such as Spanish \cite{jukiewicz_can_2026}. Addressing this gap, a key contribution of this study is the use of real-world Spanish data, combined with expert validation, within a human-in-the-loop evaluation framework.

%All in all, automated evaluation based on LLMs and well-designed rubrics and prompts has shown promising results for text-based English written evaluations. However, as aforementioned, few studies focus on Spanish texts, while it is well known that LLMs are not equally trained on different languages, leading to the fact that results from English based studies are not necessarily transferable to other languages such as Spanish texts \cite{jukiewicz_can_2026}, the use case of this study. In this regard, one of the major contributions of our work is to provide results based on Spanish real data and experts validation along with a human-in-the-loop workflow. %In this regard, \cite{} CITAR SAT? shows promising results in a real case Spanish open-ended question setting. Besides, hardly all studies point out that human oversight is needed for accuracy purposes, and needless to say, for ethical or legal reasons. 

Building on state-of-the-art models, this study designs and implements an AI-assisted evaluation system for Spanish argumentative writing in a high-stakes national exam. A prompt-engineering-based approach is developed and validated on large-scale datasets, and is integrated into a human-in-the-loop workflow that explicitly balances efficiency with expert supervision. The proposed framework is guided by a central principle: ensuring that no final decision affecting test outcomes is made without appropriate human oversight.
\added{Beyond evaluating model performance, this work therefore contributes an operational framework for integrating LLM-based scoring into a high-stakes assessment process, demonstrating how Human-in-the-Loop decision making can mitigate fairness risks while substantially reducing expert grading workload.}

%TODO ver si lo dejamos
%In this work, we build on state-of-the-art models to design and implement a real world AI-assisted evaluation system for a Spanish argumentative text evaluation in a high-stakes national exam. We develop a prompt engineering based solution for automated evaluation, which we validate against large real-world datasets, and  propose a human-in-the loop system which trades-off efficiency with human supervision, while sticking to one precept: no one is left behind without human supervision. 

 %Students value more complete and more useful detailed and personalized feedback, which comes from AI based models, while they still prefer poorer human-based ones \cite{henderson_comparing_2025}.
 
%algunos ponen foco en un approach automatico, nosotros buscamos un human in the loop ethichal one

%hablar de los diferentes enfoques para HITL

%en nuestro caso es tailored para texdto argumentativo en español, y contamos con muchos datos

\section{Context and Case Study}
%\section{Case Study: A National Writing Assessment}
\label{sec:context}
The study focuses on a knowledge accreditation test \added{called Prueba Nacional de Acreditación de la Educación Media Básica (Acredita EB)}\footnote{Prueba Acredita EB - ANEP: https://acredita.anep.edu.uy/.}\added{, organized by the National Public Education Administration (ANEP), the public authority responsible for primary and secondary education in Uruguay.}
This is a large-scale assessment that has been administered annually since 2020, with 3,604 participants in its first edition and reaching its highest number of participants in 2024 (6,204). The target population consists of individuals over 21 years of age who completed primary school but subsequently dropped out of the educational system. The test provides an opportunity to obtain lower secondary education certification, equivalent to completing the third grade of basic secondary school. The continuity of its administration over the years has enabled the compilation of a substantial volume of human-scored data, contributing to the validity and reliability of the results.

Because the test tasks vary each year and the rubric has evolved over time, only data from the two most recent editions were considered for this study. The 2024 scoring data were first used to develop an AI-based model to support the grading process, and the 2025 data were subsequently used for validation.

\subsection{Structure of the Exam}
The test consists of three sections: Reading Comprehension, Problem Solving, and Writing. It is administered individually on a computer, without access to study materials or internet resources. All sections are completed within a single test session lasting 2 hours and 50 minutes, including a 10-minute break. The sections are graded independently, with separate technical teams and examiners responsible for each of them.

The Reading Comprehension section evaluates the ability to understand, interpret, and critically analyze written texts. It includes locating explicit information, inferring global themes and implicit meanings, identifying the author’s intentions, and critically reflecting on texts in relation to social and communicative contexts. In addition, it assesses linguistic awareness, such as recognizing and interpreting the use of punctuation, connectors, graphic resources, and the meanings of words and expressions—both literal and figurative—within a text.

The Problem Solving section assesses individuals’ ability to understand and characterize problems, design and implement appropriate solution strategies, and evaluate and communicate results. This includes identifying and organizing relevant information, relating variables and making inferences, applying knowledge and basic procedures, evaluating the quality of information sources and solutions, and clearly communicating strategies, reasoning, and evidence-based conclusions.

In the Writing section, candidates are asked to produce an argumentative text on a specific topic. Each year, a different theme is defined, addressing different types of issues, such as environmental–ecological, ethical–social, and other related topics. Participants are expected to develop a clear position, provide supporting arguments, and demonstrate control of text organization, cohesion, and language conventions.

The first two sections consist primarily of multiple-choice questions and are therefore scored automatically. In contrast, the written production is evaluated manually using a detailed rubric and a procedure established by a specialized language team. As a result, the evaluation of the Writing section is the most time- and resource-intensive component of the overall scoring process, and applicants typically receive their final results no sooner than three months after the examination.

%\subsection{Human evaluation procedure}
\subsection{Evaluation Procedure}
\label{subsec:human_evaluation_process}

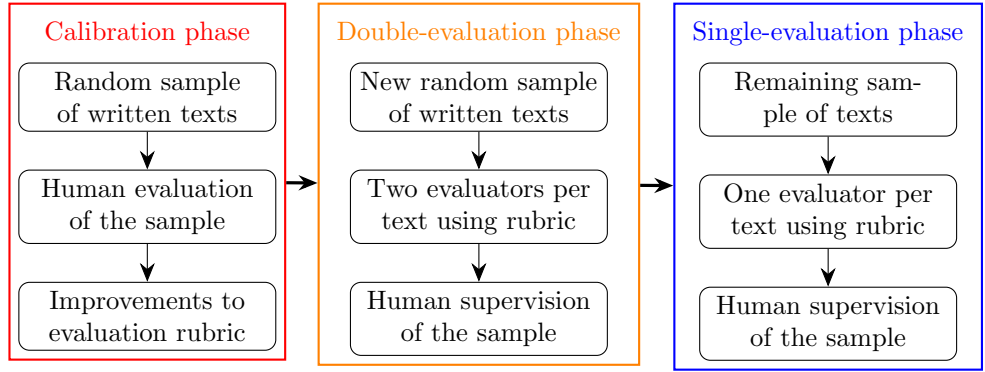
\begin{figure}
    \centering
    \tikzstyle{block} = [rectangle, draw, fill=blue!20, text width=9em, text centered, rounded corners, minimum height=2em]
     \tikzstyle{line} = [draw, -{Stealth[length=2.5mm, width=2mm]}]
     \tikzstyle{group} = [font=\fontsize{6}{6}\color{black}]
     	
     \begin{tikzpicture}[node distance = 1.5cm, auto]
     	% Place nodes
        \node [group, color=red] (first_phase) {Calibration phase};
     	\node [block, fill=white, below = 0.1cm of first_phase] (init) {Random sample of written texts};
      	\node [block, fill=white, below = 0.5cm of init] (development) {Human evaluation of the sample};
     	\node [block, fill=white, below = 0.5cm of development] (2025) {Improvements to evaluation rubric};
        \node [group, color=orange, right = 0.9cm of first_phase] (second_phase) {Double-evaluation phase};
     	\node [block, fill=white, below = 0.1cm of second_phase] (second_sample) {New random sample of written texts};
        \node [block, fill=white, below = 0.5cm of second_sample] (second_evaluation) {Two evaluators per text using rubric};
        \node [block, fill=white, below = 0.5cm of second_evaluation] (second_supervision) {Human supervision of the sample};
        \node [group, color=blue, right = 0.7cm of second_phase] (third_phase) {Single-evaluation phase};
        \node [block, fill=white, below = 0.1cm of third_phase] (third_sample) {Remaining sample of texts};
        \node [block, fill=white, below = 0.5cm of third_sample] (third_evaluation) {One evaluator per text using rubric};
        \node [block, fill=white, below = 0.5cm of third_evaluation] (third_supervision) {Human supervision of the sample};

        %\node [block, fill=white, below = 0.5cm of second_supervision] (TRI) { Item Response Theory};
        %\node [block, fill=white, below = 0.5cm of TRI] (Bookmark) {Bookmark method};
        %\node [block, fill=white, below = 0.5cm of Bookmark] (Final) {Evaluation level classification};

        \node [draw=black, color=red, thick, fit={(first_phase) (init) (development) (2025)}] (phase1) {};
        \node [draw=black, color=orange, thick, fit={(second_phase) (second_sample) (second_evaluation) (second_supervision)}] (phase2) {};
        \node [draw=black, color=blue, thick, fit={(third_phase) (third_sample) (third_evaluation) (third_supervision)}] (phase3) {};
     	
     	% Draw edges
     	\path [line] (init) -- (development);
     	\path [line] (development) -- (2025);
        \path [line, line width=1.2pt] (phase1) -- (phase2);
        \path [line] (second_sample) -- (second_evaluation);
        \path [line] (second_evaluation) -- (second_supervision);
        \path [line, line width=1.2pt] (phase2) -- (phase3);
        \path [line] (third_sample) -- (third_evaluation);
        \path [line] (third_evaluation) -- (third_supervision);
        %\path [line, line width=1.2pt] (phase3) |- (TRI);
        %\path [line] (TRI) -- (Bookmark);
        %\path [line] (Bookmark) -- (Final);

 	\end{tikzpicture}
 	\caption{Evaluation workflow for the Writing section.}
    \label{fig:human-process}
 \end{figure}

Since the multiple-choice sections are scored automatically, most of the human effort in the test evaluation process is devoted to the Writing section. Two different groups of professionals participate in this process: the language assessment expert team and the raters. The language experts define the evaluation rubric, which is described in detail in the next subsection. They also train and supervise the raters’ application of the rubric. Figure~\ref{fig:human-process} illustrates the current evaluation workflow for the Writing section.

The first stage, known as the calibration phase, involves the expert team rating a random set of written texts (typically around 50), with each response evaluated by multiple raters. This stage concludes with discussion meetings in which the coding of each response is reviewed in order to reach a consensus score. The outcome of this stage may lead to adjustments to the rubric. In Figure~\ref{fig:human-process}, this phase corresponds to the blocks grouped within the red frame.

In the second stage, based on a new and larger random sample of written productions, a double-scoring procedure is conducted. Each text is scored by two raters and supervised by a member of the expert team. The goal of this stage is to measure inter-rater agreement in order to verify that the rubric and its application are sufficiently clear and reliable. This phase is highlighted within an orange box in Figure~\ref{fig:human-process}.

In the third stage, shown within a blue box in Figure~\ref{fig:human-process}, the remaining written responses are evaluated by a single rater. At this stage, a random sample of responses is selected for expert team review. By the end of this stage, all written responses have been assigned scores.

After the evaluation process is complete, each of the three test sections receives a numerical score for every participant. Next, the same methodology is applied independently to each section in order to determine the sufficiency levels. An Item Response Theory (IRT) model \cite{hambleton_fundamentals_1991,vanDerLinden2016IRT,Chen2025IRTFramework} is used to estimate the latent variable representing each student’s ability based on their responses. The bookmark method \cite{Karantonis_2006, mitzel_bookmark} is then applied, combining the statistical calibration from the IRT model with expert judgment to define the cut scores that establish three possible performance levels for each test section: \textit{Proficient}, \textit{Close to Proficiency}, and \textit{Insufficient}. 

%Finally, once all written responses have been assigned scores, an Item Response Theory (IRT) model is applied to estimate the underlying variable that reflects each student’s ability based on their responses. The bookmark method is then used, combining the statistical calibration from the IRT model with expert judgment regarding the valuation of the different rubric items, to define the cut scores that establish the three possible performance levels: insufficient, close to proficiency, and proficient.

Finally, based on the resulting proficiency levels across the test sections, a pass/fail decision is made for each student.
In order to pass the test, participants must obtain at least two sections rated as \textit{Proficient}, and the remaining section must be rated as either \textit{Proficient} or \textit{Close to Proficiency}.

\subsection{Writing Evaluation Rubric}

%As previously mentioned, for the Writing section evaluation each year a marking rubric is defined by the team of language experts. The rubric for the last two test editions (2024 and 2025) consists of 15 different items, most of them binary, which are grouped into three domains:

As previously mentioned, the evaluation of the Writing section each year is based on a scoring rubric defined by a team of language assessment experts. This rubric constitutes one of the most important technical components of the assessment. It is organized into three complementary domains: discursive, textual, and orthographic. The first focuses on content and communicative intention, the second on organization and textual flow, and the third on technical correctness. Each domain groups specific indicators that assess key aspects of written production, assigning scores according to predefined criteria. The evidence associated with each indicator follows differentiated coding schemes, which may take the form of either dichotomous criteria or progressive scales depending on the type of performance being evaluated. Due to space limitations \added{and confidentiality restrictions associated with the operational assessment}, the full rubric and its application manual are not included in this article; however, the domains and their respective items are described below.

\paragraph{Discursive Domain}
%The discursive domain considers the characteristics of the communicative situation proposed by the test, how information is organized in the written response, and the register and vocabulary used. In this domain, six items are evaluated. These include whether the text expresses a clear opinion and adopts an argumentative stance, whether it presents an identifiable introduction and conclusion, and the appropriateness of its register and vocabulary.
The discursive domain refers to the ability of the text to fulfill its communicative purpose, to be appropriately structured, and to employ language suitable for an academic context. It evaluates how the student addresses the task, organizes ideas, and uses language to communicate a clear and coherent message. This domain includes three areas, each of which contains two rubric items.
Communicative adequacy is evaluated by verifying whether the text expresses an opinion and whether it presents a defined position supported by arguments. The structure of the text is assessed through two items that verify whether it is properly organized, including an appropriate introduction and conclusion. Finally, language use is evaluated through the items register and vocabulary. Register is assessed according to whether the text is appropriate for the communicative situation, without markers of informality or orality. Vocabulary evaluation examines whether the words used are appropriate for the context and sentence structure, without usage errors or unnecessary repetitions that could affect fluency. All items in this domain use binary scoring values, except for argumentation, which has three levels of proficiency.

\paragraph{Textual Domain}
%The textual domain refers to all elements of the internal structure of the text. It is evaluated in seven items: coherence, thematic progression, paragraph structure, textual cohesion, use of pronouns, use of connectors, agreement (nominal and verbal), syntax, and sentence construction.
The textual domain focuses on the use of various grammatical mechanisms at both the text and sentence levels. These mechanisms contribute to the overall unity of the text, either by maintaining referential continuity or by preserving the relationships between different ideas. This domain is organized into four areas comprising seven scoring items.
Coherence is evaluated through two items: thematic progression and paragraph structure, highlighting the importance of a clear organization that facilitates reading. Two additional items correspond to the evaluation of textual cohesion: the use of connectors or linking expressions, and the use of pronominal references. The rubric also evaluates grammatical agreement, both nominal—between nouns and their determiners, or between nouns and their attributes or predicatives—and verbal, that is, between subject and verb. Finally, the last item in this domain corresponds to syntax, which evaluates the structural completeness and coherence of sentences. All items in this domain use binary scoring values.

\paragraph{Orthographic Domain}
%The orthographic domain, in turn, assesses mastery of the rules specific to written communication. It is structured around two indicators: punctuation and spelling.
The orthographic domain evaluates the correct application of the conventions governing written language, including punctuation, accentuation, and spelling, as well as the overall formal presentation of the written production. This domain focuses on identifying errors that compromise legibility and affect the effective communication of the message. The evaluation considers both the absence and the incorrect use of orthographic signs and rules.
This domain is coded using two items: punctuation and spelling. Punctuation is evaluated on a scale from 0 to 2, where 0 corresponds to three or more errors, 1 corresponds to one or two errors, and 2 corresponds to the absence of punctuation errors. For spelling, the number of errors is counted (up to a maximum of 10) and divided into quartiles to assign levels of sufficiency, resulting in four possible performance levels.

\paragraph{Rubric Summary}
%\textcolor{red}{Sacaría esto porque no suma. Lo anterior me parece espectacular!}\sout{Finally, it is important to highlight that the emphasis on topic relevance, argumentation, and personal reflection indicates that the task aims not only to assess technical language proficiency but also critical thinking and personal engagement with the content. At the same time, the detailed attention given to elements such as agreement, connectors, and punctuation reflects the expectation of a carefully written and professional text aligned with high academic standards.}
In summary, the 15 rubric items are: Opinion, Argumentation, Introduction, Conclusion, Register, Vocabulary, Thematic progression, Paragraph structure, Connector use, Pronominal references, Nominal agreement, Verbal agreement, Syntax, Punctuation, and Spelling.
The preceding description of each domain and its respective items illustrates the technical complexity involved in interpreting the rubric for scoring purposes. This complexity is an important aspect to consider when attempting to replicate the task using a language model and when designing the corresponding prompts.
\added{Finally, it is important to note that the writing evaluation rubric does not assign predefined weights to its 15 items. Instead, proficiency levels are determined as previously mentioned, by applying Item Response Theory (IRT) to estimate the psychometric properties of the items, followed by the Bookmark method to establish the cut scores associated with each proficiency level.}

\section{Methodology}
%\section{AI-Assisted Scoring Methodology}
\label{sec:methodology}
% Posible titulo Research Design
In this section, we describe the methodology used to develop the AI-based model for evaluating written responses. We begin with an overview of the dataset and then present an illustrative example of the general prompt structure.

%\subsection{Dataset Preparation and Preprocessing}
\subsection{Dataset Description}

The dataset considered in this study corresponds to the last two editions of the test (2024 and 2025). For each edition, we have the proficiency levels achieved by each participant in each test section (Reading Comprehension, Problem Solving, and Writing). Furthermore, for the Writing section, the dataset includes the written responses produced by each participant and the item-level scores assigned by the raters for each rubric criterion.

For the 50 texts used in the calibration phase, we also have the item-level scores assigned independently by each member of the language expert team, as well as the final consensus scores reached after discussion. For the remaining texts, only a single score assigned by a rater is available; this score represents the final result, validated through random sample supervision by the team of language experts.

\begin{figure}
  \centering
  \includegraphics[width=\textwidth]{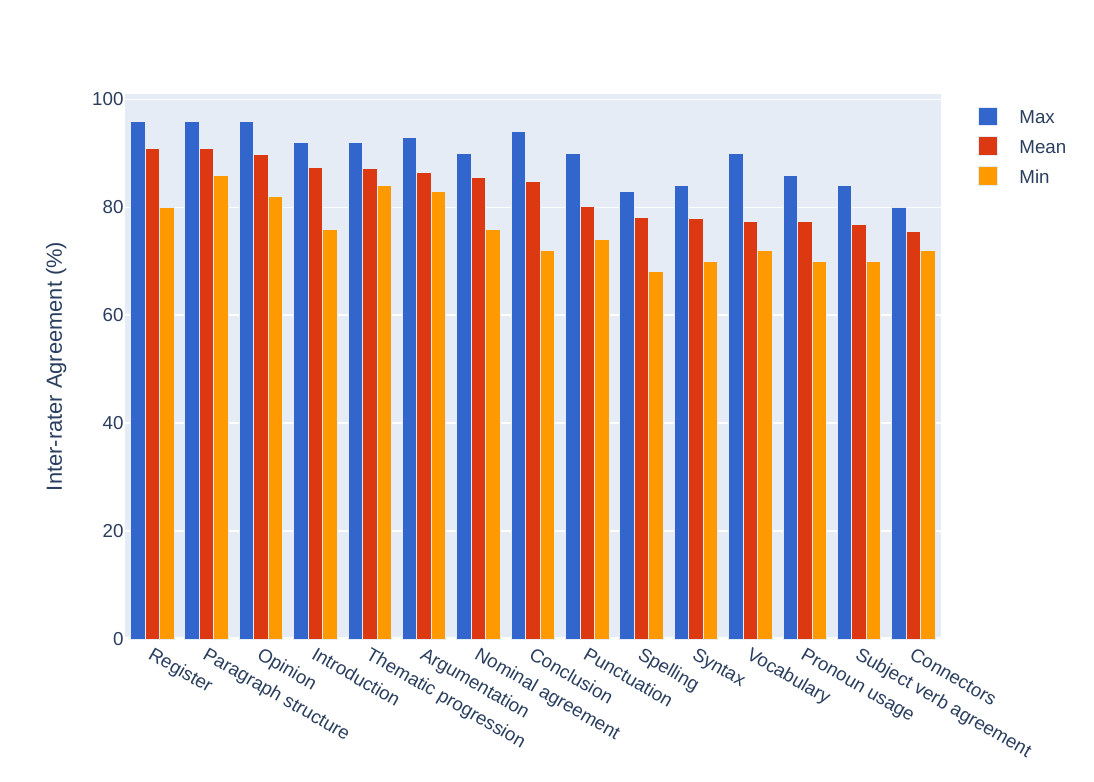}
  \caption{Inter-rater agreement analysis in the 2024 Calibration Phase.}
  \label{fig:calibration_phase_agreement}
\end{figure}

For the purposes of this study, the available data allow us to analyze agreement among human evaluators during the calibration stage.
In 2024, ten evaluators participated in this process, where they all separately corrected each of the 50 text productions in the calibration sample.
This enables us to establish an empirical upper bound on the performance that could reasonably be expected from an AI-based model. Figure~\ref{fig:calibration_phase_agreement} presents the results of this analysis for the 2024 calibration phase. For each rubric item, the figure reports the minimum, average, and maximum agreement between individual raters and the consensus score reached after expert discussion.

It can be observed that there is no item for which all evaluators agree on the final assigned score, and that some items show greater discrepancies than others.
The results show that the minimum value, that is, the rater exhibiting the greatest discrepancy with the consensus score, generally achieves agreement below 80\%. In contrast, the maximum agreement exceeds 90\% for several rubric items, although for others it remains closer to 80\%. It is usually considered that the level of agreement between raters for this type of test should be at least 70\% to be acceptable.

\added{
To complement the percentage agreement analysis, we also computed Cohen's Kappa coefficient for each rubric item, providing a chance-corrected measure of inter-rater agreement. According to the interpretation scale proposed by~\cite{Landis1977}, more than half of the rubric items achieved either \emph{substantial} agreement (Introduction, Register, Paragraph structure, and Nominal agreement) or \emph{moderate} agreement (Conclusion, Pronoun usage, Subject--verb agreement, Syntax, and Punctuation). Two additional items, Vocabulary and Thematic progression, showed \emph{slight} agreement. Finally, Opinion, Argumentation, and Connectors exhibited \emph{fair} agreement. Inspection of these latter cases revealed that the relatively low Kappa values are largely explained by the skewed distribution of scores, as the vast majority of responses satisfy these rubric criteria, reducing the coefficient despite the high observed agreement among raters.
}

\replaced{
Finally, considering the operational human scores as the reference labels for this study, we are able to evaluate the performance of the AI-based model. It is important to note that multiple independent human ratings are only available for the calibration sample discussed in the previous section. For the remaining responses, a single operational score assigned by an expert rater is available, which is the score used in the official assessment process and is therefore adopted as the reference label throughout this study. As the inter-rater agreement analysis showed, some level of disagreement between equally qualified human raters is expected, implying that these reference labels may contain a limited number of inconsistencies. Nevertheless, they represent the only labels available for the complete dataset and therefore constitute the operational ground truth against which the AI model is evaluated. For this purpose, a random sample of 1,000 written responses was selected. This procedure was applied to both the 2024 and 2025 datasets. In both cases, the data used for performance evaluation were not used to refine the corresponding prompts. The selected sample size ensures the statistical validity of the performance estimates while avoiding the increased costs associated with running the AI model on the entire dataset.
}{
Finally, considering the scores assigned by the raters as ground truth, we are able to evaluate the performance of the AI-based model. For this purpose, a random sample of 1,000 written responses was selected. This procedure was applied to both the 2024 and 2025 datasets. In both cases, the data used for performance evaluation were not used to refine the corresponding prompts. The selected sample size ensures the statistical validity of the performance estimates while avoiding the increased costs associated with running the AI model on the entire dataset.
}

%The dataset includes the written responses produced by each participant and the score assigned to each rubric item. For the 50 texts used in the calibration phase, we also have the item-level scores given independently by each evaluator, as well as the final consensus score reached after discussion. For the remaining texts, only the final agreed-upon score, validated through supervision, is available. This format enables the evaluation of each rubric item for every text using AI and to assess the accuracy with which its outputs align with human raters’ evaluations.

%A random sample of written texts from the 2024 and 2025 editions was selected. Each text was evaluated using a 15-item rubric; however, an item was excluded from the AI-based evaluation and was scored without the use of an LLM. Consequently, only 14 items were used in the comparison with model outputs. The human evaluation labels served as the ground truth against which the model outputs were compared. A random sample of 1000 written texts was selected from approximately 5000 submissions in each of the 2024 and 2025 editions.

\subsection{Prompt Engineering}

The strategy used to develop the AI-based model consisted of operationalizing each rubric item through dedicated prompts that assign a score (generally binary) to each written response, identify the most relevant text excerpt, and provide a justification for the assigned score.
The original dataset does not include evidence supporting the score assigned to each rubric item. However, generating such evidence is considered useful for supporting and explaining the scores produced through AI-based evaluation.

\replaced{To choose the model we only had two possible options: open models running locally but with reduced computing capacity, or the use of an institutional licensed account to run proprietary models. 
After preliminary tests, the selected LLM was the OpenAI GPT-5 model, which was the most advanced reasoning model from OpenAI at the time of conducting the study.
The default values were used for the parameter configuration, with the reasoning effort level set to Medium. Additionally, the structured output functionality was used to ensure the JSON format of the model's output.}{After preliminary testing with different models, the selected system was the OpenAI GPT-5 model with a standard parameter configuration (e.g., reasoning effort set to Medium), accessed via API calls using an institutional licensed account.}

A prompt-based strategy was implemented: for each rubric item, a specific prompt was designed in Spanish, based on the rubric criteria and the detailed description of how the item should be evaluated. Each prompt was initially tested on a sample of 10 texts. This iterative process made it possible to identify cases not explicitly covered in the rubric description but implicitly considered by human raters during manual scoring.

After refinement, the system was applied to a sample of 1,000 written responses (approximately one fifth of the total dataset) to estimate the agreement between AI-based scoring and the final human-assigned scores for each response. Human evaluation was treated as the ground truth, as the objective was to approximate as closely as possible the assessments performed by expert human raters.

The final prompts follow the structure outlined below:
\begin{quote}
\begin{itemize}
    \item[$-$] \textbf{Introduction.} E.g.: ``Correct errors of ITEM in an argumentative text, strictly according to the rubric detailed below. It is not allowed to assume rules that are not explicitly stated here. Definitions related to item.''
    \item[$-$] \textbf{Item description.} E.g.: ``What is evaluated? Agreement between the noun and its determiners, or the noun and its predicative.''
    \item[$-$] \textbf{Evaluation Rules.} Rules applied by the evaluator to penalized or not the text using the item description.
    \item[$-$] \textbf{Explanation of Codes.} ``Assign Code 1 when ..." and "Assign Code 0 when ...''
    \item[$-$] \textbf{List of examples.} Correct Example (Code 1) and Incorrect Example (Code 0).
    \item[$-$] \textbf{Output Format.} E.g.: ``Output Format (maximum 8 tokens in fragment and explanation, maximum 3 errors; if Code 1 → ``errors'': [])'':
    \begin{verbatim}
    {
      "code": "0",
      "errors": [
        {"fragment": 'este crianza puede verse', 
        "explanation": 'Determinante ("este") no concuerda 
        con crianza'}]
    }
    \end{verbatim}
\end{itemize}
\end{quote}

%\added{A full prompt example is included in Appendix~\ref{ap:promptexample}.}

\subsection{Evaluation Metrics}

The model outputs were compared with human scores at both the item level and the aggregate level. Accuracy was computed as the proportion of exact matches between the model predictions and the final human evaluations.

In addition, we conducted a cross-year validation to assess prompt generalization. Specifically, models were tested on the 2025 responses using rubric prompts and contextual patterns derived from the 2024 data, without any retraining. This procedure provides insight into the robustness of the approach across exam editions.

\section{Results}\label{sec:results}
This section presents the results of the experiments conducted with the proposed AI-based model. First, we analyze its performance during the calibration phase of the 2024 test and evaluate the consistency of its outputs (i.e., whether the AI model assigns the same score consistently). Next, we examine how this performance generalizes to a large sample from the 2024 test. We then assess whether the model generalizes to data from the following year’s test (2025). Finally, we analyze how the AI model’s results affect the decision-making process regarding writing proficiency levels and, ultimately, the pass/fail outcomes.

\subsection{Calibration AI-model Accuracy and Consistency Analysis}
The first analysis focuses on the 2024 calibration phase. Based on these data, minor adjustments were made to the prompts derived from the evaluation rubric. Therefore, these results can be interpreted as reflecting the model’s performance during a development stage, since the same data used for evaluation also informed prompt refinement. Furthermore, each of the 50 texts in the calibration phase was evaluated 10 times by the AI model to assess output consistency.

\begin{figure}
  \centering
  \includegraphics[width=\textwidth]{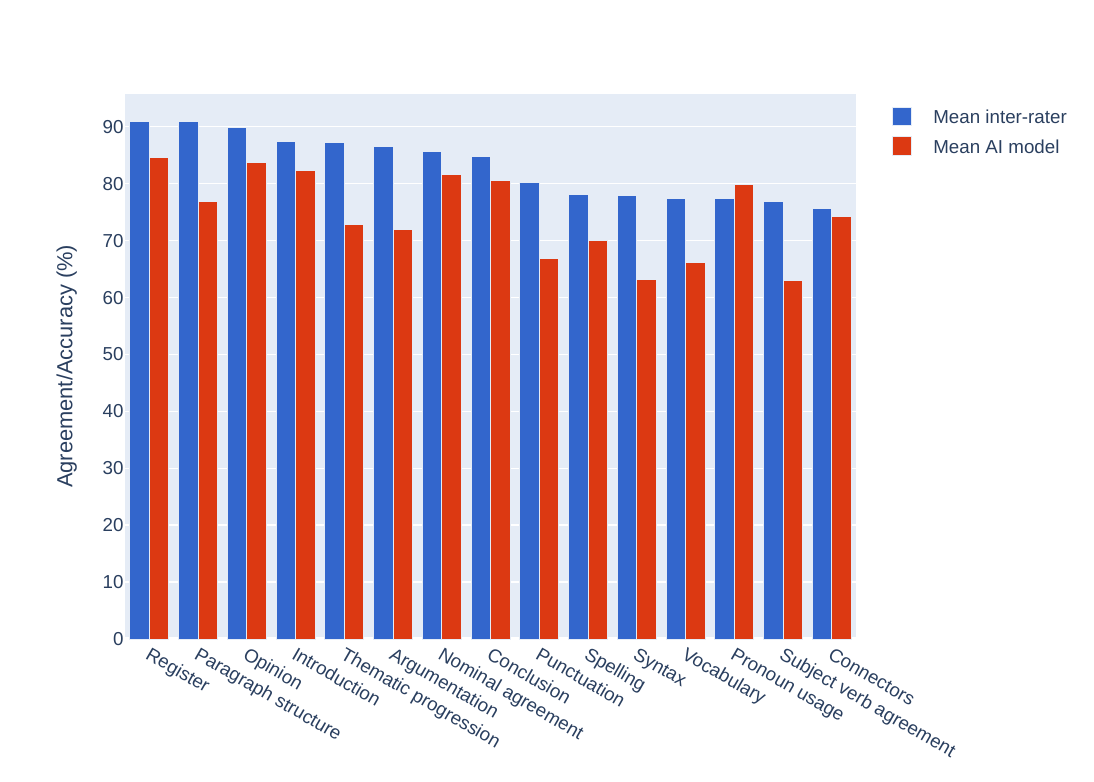}
  \caption{AI model accuracy vs. average inter-rater agreement in the 2024 Calibration Phase.}
  \label{fig:calibration_phase_AI}
\end{figure}

Figure~\ref{fig:calibration_phase_AI} reports the average AI model accuracy across the 10 runs. For reference, the average inter-rater agreement is also shown for comparison. As expected, the model’s results fall slightly below the agreement observed between human raters. However, for more than half of the rubric items, this difference is approximately 5\%, and for the remaining items, it never exceeds 15\%. In absolute terms, AI model accuracy ranges between 60\% and 80\%.

\begin{figure}
  \centering
  \includegraphics[width=\textwidth]{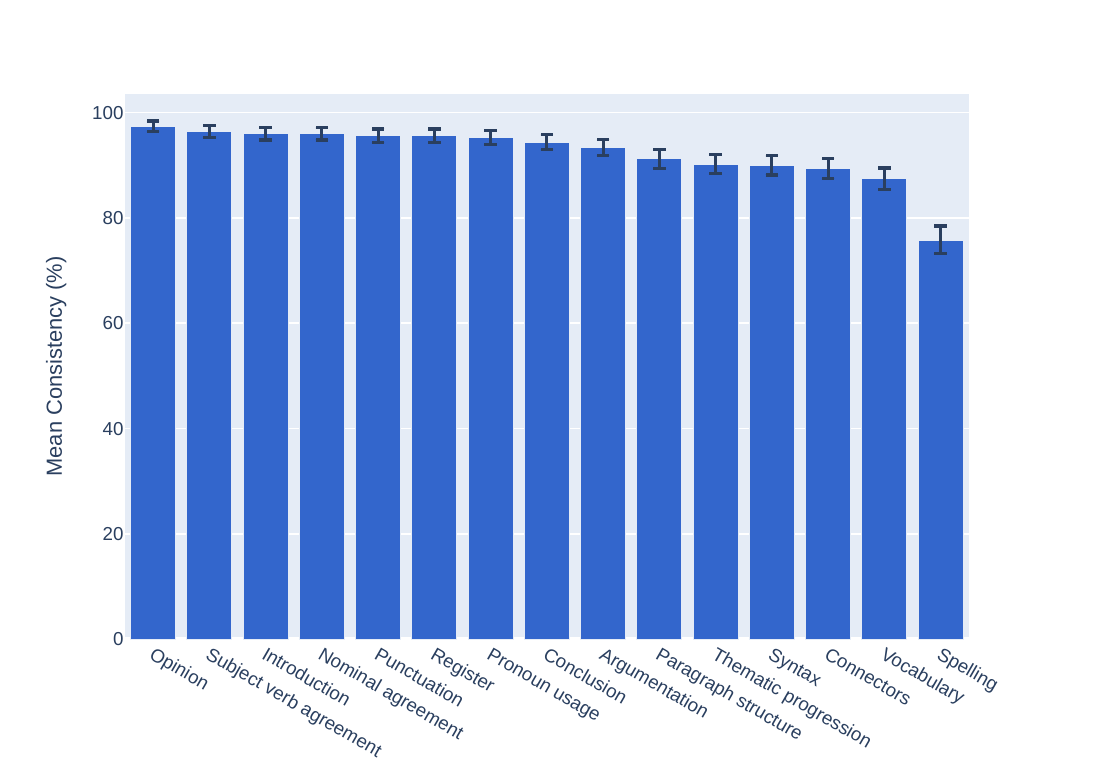}
  \caption{AI model consistency analysis in the 2024 Calibration Phase. \added{Error bars indicate 95\% confidence intervals.}}
  \label{fig:AI_consistency}
\end{figure}

Regarding the AI-model output consistency, the results are presented in Figure~\ref{fig:AI_consistency}. For the vast majority of rubric items, consistency approaches or exceeds 90\%. This indicates that, on average, the same score is assigned in 9 out of 10 repeated evaluations, reflecting a high degree of stability. It is important to note that the AI model’s output is inherently non-deterministic, which makes consistency analysis particularly relevant.

The only item that shows substantially lower consistency corresponds to spelling, which aligns with the typical behavior of large language models.
For this reason, the use of the AI model for this rubric item was replaced with a deterministic NLP-based approach. The selected method was LanguageTool \cite{Naber2003}, an open-source grammar checker that enables the detection of grammatical and spelling errors. A filter was also applied to the LanguageTool output to remove corrections that the rubric explicitly assigns to other items rather than to spelling \added{(e.g. nominal or subject-ver agreement)}.
This approach was validated with the calibration sample, where a comparable performance was achieved with that obtained with LLMs (68\%), but with the advantage that this approach, being deterministic, all executions give the same result (its consistency is 100\%).
In the following results, this method is used to score the spelling-related rubric item, while the remaining items are evaluated using the developed AI model.

%For 2025, the number of evaluators who participated in this phase was smaller, with four evaluators assessing 50 texts. The Figure~\ref{fig:calibration_phase_agreement} (right) shows the level of agreement with the final assigned score among these evaluators for each item. In this year, it can be observed that, with fewer evaluators, there are items with 100\% agreement.

%\subsection{Accuracy Human-AI random sample (1000 texts)}
\subsection{Large-scale AI model performance for 2024 and 2025 tests}

\begin{figure}[!ht]
    \centering
    \includegraphics[width=\textwidth]{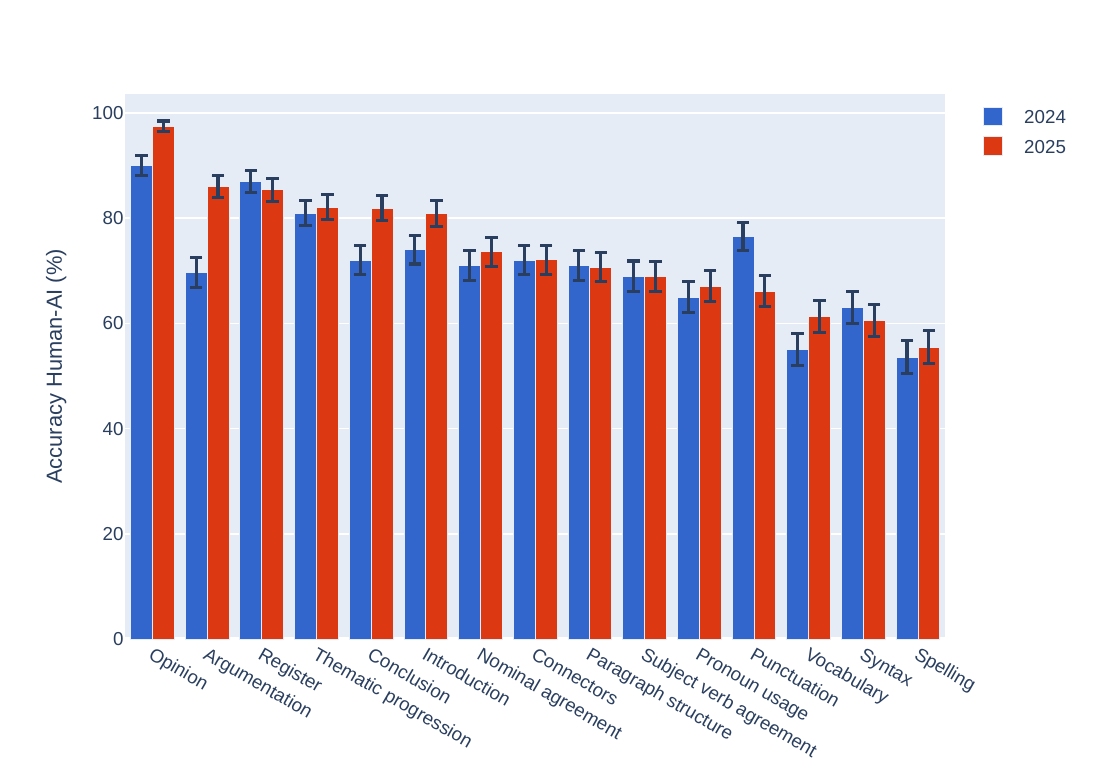}
    \caption{Human--AI agreement by rubric item for random samples from the 2024 and 2025 assessments. \added{Error bars indicate 95\% confidence intervals.}}
    \label{fig:sample_agreement}
\end{figure}

After the evaluation using the calibration-phase data, the first challenge was to determine whether the model maintained its performance in a large-scale evaluation using data that had not been used during the prompt adjustment stage. To this end, a random sample of 1,000 texts from the 2024 test writing section was selected, and the model's accuracy was analyzed using the scores assigned by human raters as ground truth. Figure~\ref{fig:sample_agreement} shows the results. The most notable finding is that the model's performance remained similar to that observed in the calibration phase, with accuracy ranging between 60\% and 80\% for most rubric items. The items with the lowest scores are those related to vocabulary, syntax, and spelling. This can be explained by the relative complexity of the rubric in specifying what is penalized and what is not—distinctions that are difficult for an LLM operating at the token level to capture.

The next step was to extend the evaluation to data from a subsequent year. For this purpose, a new sample of 1,000 texts was considered, this time from the writing section of the 2025 test. At the prompt level, only minor adjustments were made to those rubric items that explicitly refer to the writing task instructions, as the specific topic assigned to candidates changes every year.

Figure~\ref{fig:sample_agreement} presents the results, showing that the observed performance variations with respect to 2024 were very small. This is a relevant finding, as it indicates that prompts developed for one year’s assessment can generalize to another year’s test without major effort, assuming a similar level of agreement with human raters.

\added{
To complement the agreement analysis, Cohen's Kappa coefficients were also computed for the 2025 evaluation sample, providing a chance-corrected measure of agreement between the AI model and the operational human scores. Using the same interpretation scale proposed by~\cite{Landis1977} as in the human inter-rater analysis, the agreement levels were, as expected, lower than those observed between human evaluators during the calibration phase. Most rubric items achieved either \emph{moderate} agreement (Introduction, Conclusion, Register, Nominal agreement, and Subject--verb agreement) or \emph{fair} agreement (Opinion, Paragraph structure, Pronoun usage, Syntax, and Punctuation). The remaining items (Argumentation, Vocabulary, Thematic progression, and Connectors) showed \emph{slight} agreement. Similar to the human inter-rater analysis, some of these lower Kappa values are explained by the skewed distribution of scores for rubric items that are satisfied by the vast majority of responses, while others correspond to the rubric dimensions exhibiting the lowest agreement throughout the study. Overall, the results are consistent with the expected reduction in agreement when comparing AI-based evaluations against operational human scores instead of comparisons among expert human raters.
}

\subsection{Accuracy on Proficiency Levels and Pass/Fail Decision}

As described in Section~\ref{subsec:human_evaluation_process}, the final score assigned to each written production does not depend directly on individual item scores, but rather on cut scores established after applying Item Response Theory and the Bookmark method. Accordingly, to approximate the final evaluation of the texts, this process was replicated in an automated manner by first applying IRT and then defining cut scores based on the number of correctly answered items.
\added{Specifically, the automatically derived cut scores on the IRT scale correspond to a 67\% probability of correctly answering at least seven rubric items for the \textit{Close to proficiency} level and ten rubric items for the \textit{Proficient} level. These thresholds are very close to the operational cut scores established by expert raters through the Bookmark standard-setting procedure, indicating that the simplified automated approach provides a good approximation of the official proficiency-level classification while enabling a consistent comparison between AI- and human-generated rubric scores.
}

\begin{figure}
    \centering
    \subfloat[Level confusion matrix 2024]{%
        \includegraphics[width=0.49\textwidth]{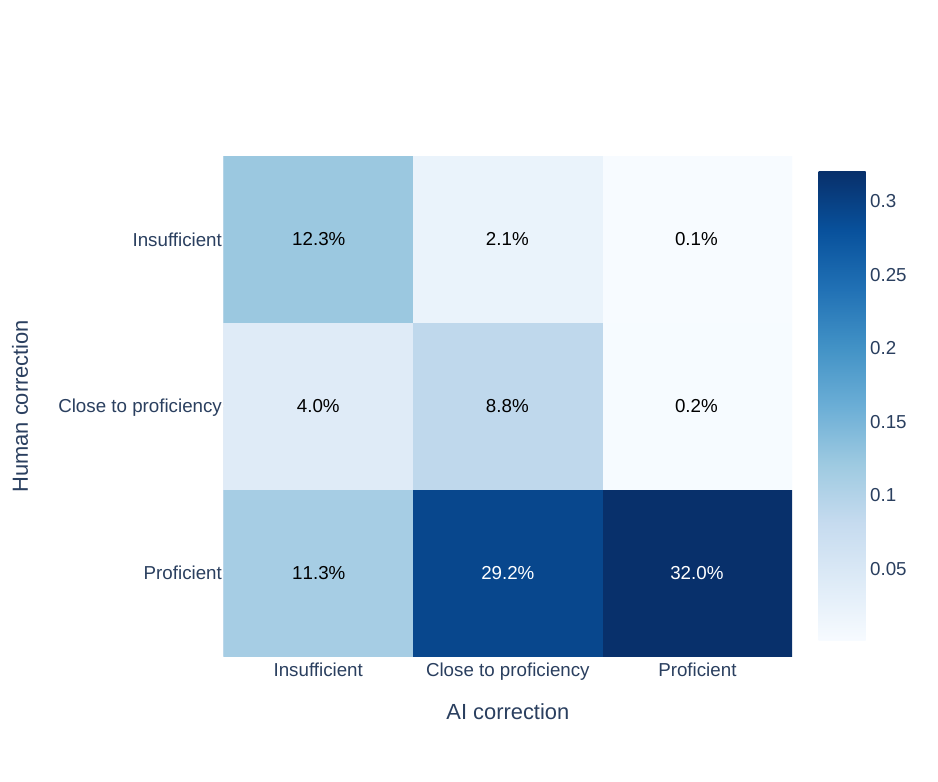}
    }
    \hfill
    \subfloat[Level confusion matrix 2025]{%
        \includegraphics[width=0.49\textwidth]{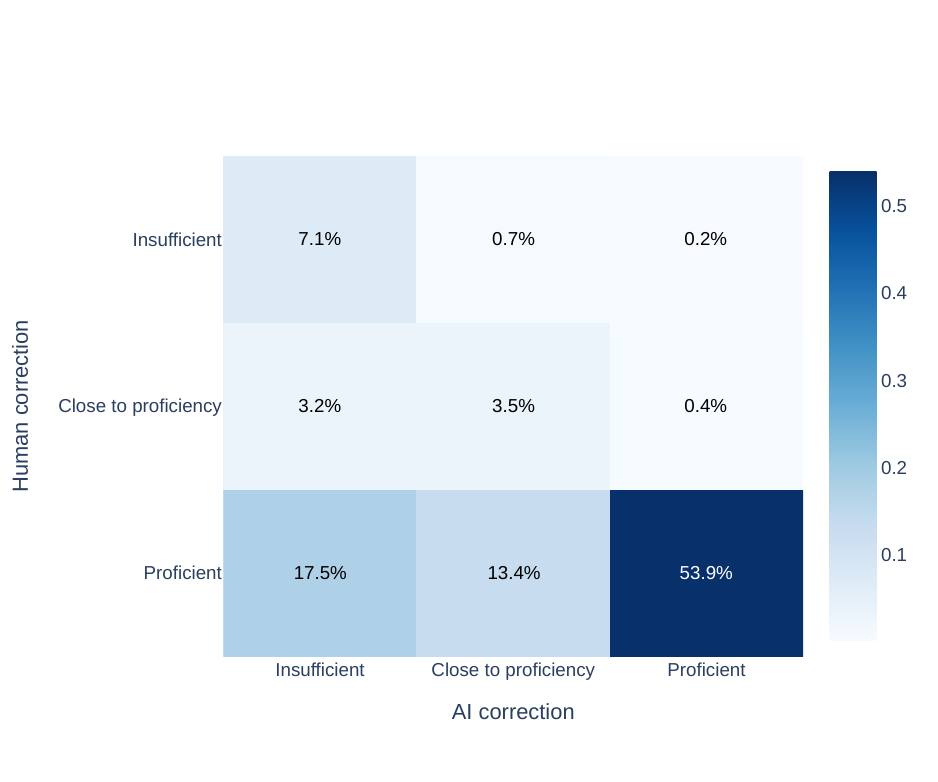}
    }
    \caption{Agreement between AI-based and human scoring across performance levels}
    \label{fig:confusion_matrix_levels}
\end{figure}

\replaced{
The same automated procedure was applied to both the AI-generated rubric scores and the human scoring results, ensuring that proficiency levels were assigned using identical cut-score criteria. This provides a fair and consistent basis for comparing the two evaluation approaches, independently of the original operational scoring workflow.
}{Although this procedure differs from the one currently used in operational scoring, it was applied to both the AI model outputs and the human scoring results, ensuring that agreement was assessed using a common criterion for performance level assignment.}
Figure~\ref{fig:confusion_matrix_levels} presents the confusion matrices for 2024 and 2025. In both cases, the AI model tends to penalize responses more strictly, indicating a consistent and predictable bias rather than random error.
This systematic behavior suggests that AI-based scoring could effectively support the overall evaluation process
\added{when integrated into an appropriate Human-in-the-Loop workflow, as discussed in the following section.}

\begin{figure}
    \centering
    \subfloat[Pass-fail confusion matrix 2024]{%
        \includegraphics[width=0.49\textwidth]{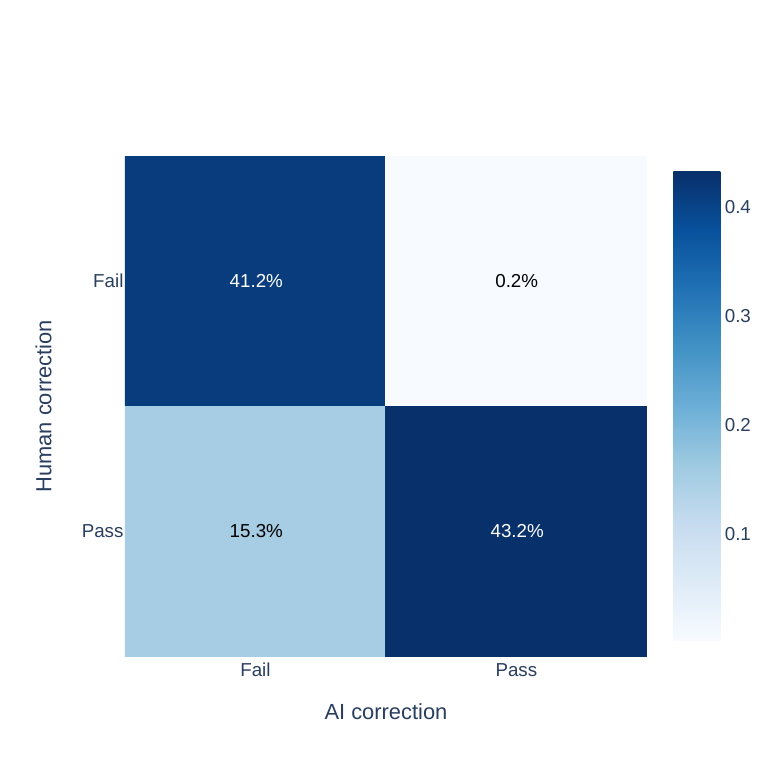}
    }
    \hfill
    \subfloat[Pass-fail confusion matrix 2025]{%
        \includegraphics[width=0.49\textwidth]{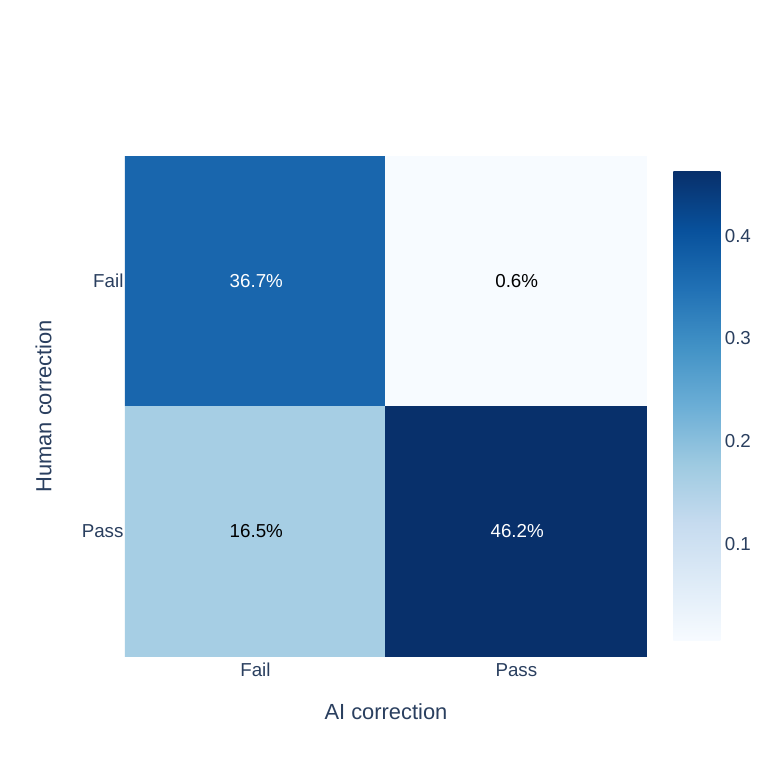}
    }
    \caption{Agreement in pass-fail decisions for the overall assessment}
    \label{fig:pass_fail_matrix}
\end{figure}

Finally, the proficiency-level data from the other sections of the test (Reading Comprehension and Problem Solving) were combined with the AI-model results for the Writing section in order to evaluate agreement in pass–fail decisions for the overall test (see Figure~\ref{fig:pass_fail_matrix}). The results again highlight the effect of the AI model’s more stringent scoring of written production, which leads to a substantial number of cases in which tests classified as passing by human raters would be classified as failing under AI-based scoring. These cases (15.3\% in 2024 and 16.5\% in 2025) are of particular interest, as they provide evidence supporting the need for human review of texts for which the AI predicts a failing outcome. 

Importantly, this bias is systematic, with very few instances observed in the opposite direction—namely, cases in which the AI model assigns a passing decision contrary to the judgment of human evaluators.
\added{A more detailed analysis by proficiency level showed that this conservative behavior is not uniform across all categories. The AI model achieved higher agreement for responses classified as \textit{Proficient} than for those classified as \textit{Insufficient}, a pattern consistent with the confusion matrices presented in Figure~\ref{fig:confusion_matrix_levels}. Consequently, the observed bias is primarily associated with under-grading rather than over-grading. While such behavior would be problematic if AI scores were used as final decisions, it provides valuable information for the design of the proposed Human-in-the-Loop workflow. In particular, responses classified as passing by the AI model can be accepted with a high degree of confidence, whereas responses classified as failing are systematically routed to expert human raters for verification before any final certification decision is made. Under this operational framework, the identified bias does not propagate to the final assessment outcome, since the cases with the greatest equity risk are precisely those subject to mandatory human review.
}

%\section{Discussion (HITL: Human in the Loop)}
\section{Human-in-the-Loop AI-Assisted Evaluation}
\label{sec:discussion}
The results obtained are encouraging and support the feasibility of incorporating AI into the scoring of written productions while maintaining quality standards and consistency with human evaluation. Although variability is observed in the LLM’s accuracy when scoring individual rubric items, often not reaching high levels of agreement with human raters, the overall tendency of the model is conservative, systematically identifying errors that are not penalized by human evaluators. While this behavior might initially appear unfavorable, when performance is examined at the proficiency-level stage and subsequently combined with the results from the other test sections, the model demonstrates acceptable overall performance.

\begin{figure}
    \centering
    \tikzstyle{block} = [rectangle, draw, fill=blue!20, text width=7.3em, text centered, rounded corners, minimum height=2em]
     \tikzstyle{line} = [draw, -{Stealth[length=2.5mm, width=2mm]}]
     \tikzstyle{decision} = [diamond, draw, fill=orange!20, text width=6em, text centered, inner sep=0.05pt]
     	
     \begin{tikzpicture}[node distance = 1.5cm, auto]\label{ams1}
     	% Place nodes
     	\node [block] (init) {Calibration phase (50-100 texts)};
      	\node [block, below = 0.5cm of init] (development) {AI ratings for all Writing responses};
     	\node [block, right = 0.3cm of development] (2025) {Reading Comprehension \& Problem Solving \\proficiency levels};
     	\node [block, below = 0.5cm of development] (IRT) {Set proficiency-levels cut scores};

        \node [decision, below=0.6cm of 2025, aspect=2] (decide) {Fails due to other test sections?};
        \node [decision, below=0.6cm of decide, aspect=2] (pass) {Is sufficient in all test sections?};
        \node [block, below =0.6cm of pass] (human_correction) {Human review of AI ratings};
        \node [block, right = 0.7cm of pass] (IRT_2) {Update proficiency-levels cut scores};
        \node [block, right =0.5cm of IRT_2] (final) {Final Writing evaluation results};
    
     	% Draw edges
     	\path [line] (init) -- (development);
     	\path [line] (development) -- (IRT);
        \path [line] (2025) -- (decide);
        \path [line] (IRT) |- (decide);
        \path [line] (decide) -- node {No} (pass);
        \path [line] (decide) -| node {Yes} (IRT_2);
        \path [line] (pass) -- node {Yes} (IRT_2);
        \path [line] (pass) -- node {No} (human_correction);
        \path [line] (human_correction) -| (IRT_2);
        \path [line] (IRT_2) -- (final);
        
 	\end{tikzpicture}
 	\caption{Proposed Human-in-the-Loop AI-assisted evaluation framework.}
    \label{fig:new_process}
 \end{figure}
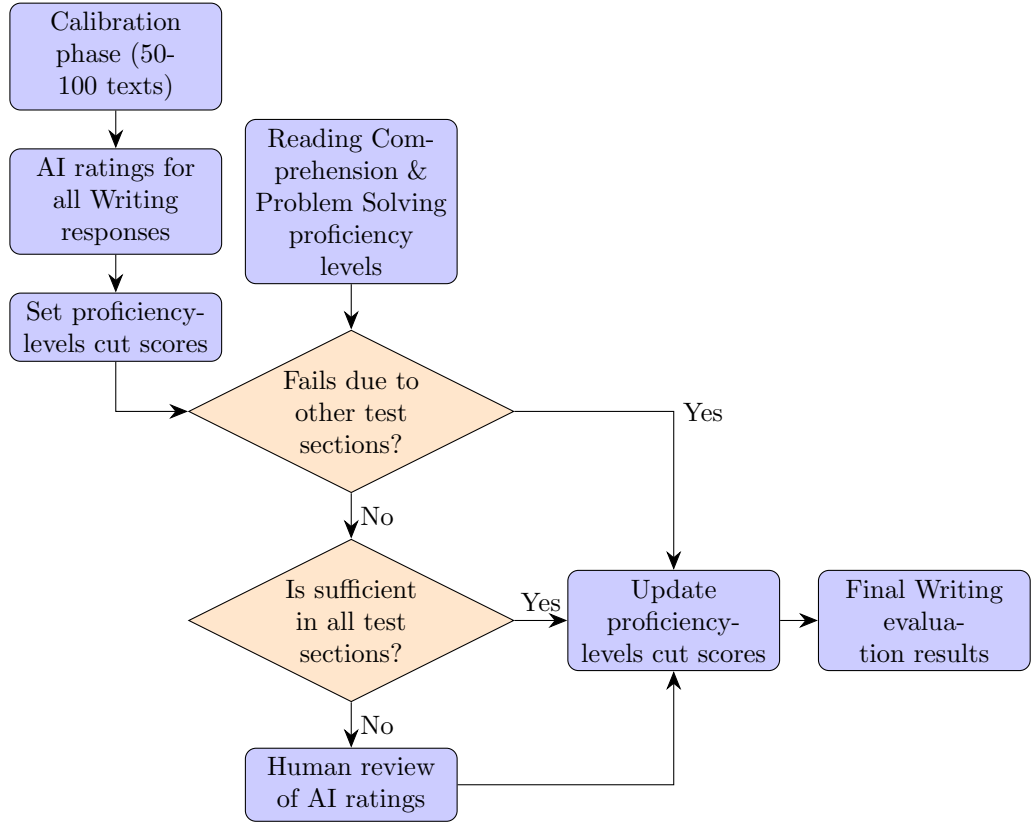

Based on these findings, a new scoring process is proposed that integrates AI-assisted Writing assessment to improve the efficiency of the overall evaluation workflow. The other sections of the test—namely Reading Comprehension and Problem Solving—are proposed to continue being scored using the current operational procedures. Figure~\ref{fig:new_process} presents a diagram of the proposed Human-in-the-Loop evaluation framework.

The first stage of the process corresponds to the calibration phase, which would be conducted following procedures similar to those used in previous editions, with the possibility of increasing the number of written productions considered (between 50 and 100). This stage would serve to validate the AI model, adapt the prompts to the specific topic of the current test, and identify potential issues arising from the calibration texts.

The second stage consists of the automated scoring of all written productions using the AI model. This step produces an initial set of item-level scores for each rubric criterion. Subsequently, IRT and the Bookmark method are applied to establish cut scores and assign the three proficiency levels defined in the scale. Candidates whose pass–fail outcome depends on the Writing section are then reviewed by human raters, supported by the AI-based scoring.

It is important to recall that, to pass the test, candidates must achieve a Proficient level in at least two sections, and the remaining section must be rated at least Close to Proficient. This implies that candidates who fail to reach at least Proficient in one section and Close to Proficient in another (considering Reading Comprehension and Problem Solving) cannot pass the test, regardless of their performance in the Writing section. Therefore, reviewing the AI-based Writing score in such cases would not affect the overall test outcome and is consequently unnecessary.

The next decision point in the workflow occurs when all three test sections are classified as Proficient. This situation arises when Reading Comprehension and Problem Solving are rated Proficient and, following AI-based scoring, the Writing section is also classified as Proficient. In this case, the candidate passes the test.
The only scenario that could reverse this outcome would be if the Writing section were actually Insufficient, since even a Close to Proficient rating would still meet the minimum requirement for passing. Therefore, the likelihood of a candidate passing due to an AI scoring error in this context is 0.2\% in 2024 and 0.6\% in 2025, given that it would require a two-level misclassification in the Writing proficiency scale. In our analyses, such cases were not observed among candidates who had achieved Proficient in the other test sections.

Finally, a recalibration of the proficiency-levels cut scores is proposed using again IRT+Bookmark. This aims to account for potential adjustments resulting from human review, which may lead to slight shifts in performance levels and, consequently, in candidates’ final outcomes.

Based on data from the most recent test editions and the AI model performance observed in this study, the proposed process would reduce by at least 50\% the number of written productions requiring full human scoring. The exact magnitude of this reduction would vary across editions, as it also depends on candidate performance in the other test sections.

\section{Conclusion and Future Directions}\label{sec:conc_fw}
This study explored the feasibility of integrating large language models into the scoring process of a large-scale standardized writing assessment. Using real operational data from the 2024 and 2025 test editions, we developed and evaluated a prompt-based AI scoring approach aligned with an existing analytic rubric and human evaluation workflow.

The results show that, although item-level agreement between the AI model and human raters varies across rubric criteria, the overall behavior of the model is stable and systematically conservative. When performance is examined at the proficiency-level stage and within the full pass--fail decision process, the AI-assisted approach achieves acceptable alignment with human-based evaluation. Cross-year validation further suggests that the proposed prompting strategy generalizes across test editions with minimal adjustments.

A key contribution of this work is the proposal of a Human-in-the-Loop (HITL) evaluation framework that strategically integrates AI-based scoring into the operational workflow. Simulation results indicate that this approach could reduce by at least 50\% the volume of written responses requiring full human scoring, while preserving decision quality in high-stakes outcomes. Beyond efficiency gains, the proposed framework maintains expert oversight in critical cases, supporting a balanced approach between automation and human judgment.

Despite these encouraging findings, several limitations and open questions remain. First, the analysis relies on agreement with human raters as ground truth, which itself exhibits non-negligible variability. Second, the study was conducted under offline experimental conditions, and operational deployment may surface additional challenges related to monitoring, drift, and edge cases.

Future work will therefore focus on multiple directions. A primary next step is the controlled pilot implementation of the proposed HITL workflow in upcoming test editions, in order to measure its impact under real operational conditions. Continued longitudinal monitoring of AI--human alignment will also be necessary to detect potential performance drift over time.

Another important research line concerns human--AI interaction effects. In particular, it will be important to study whether prior exposure to AI-generated scores influences human raters’ judgments, potentially introducing anchoring or automation bias. Experimental studies with blinded and non-blinded rating conditions could help quantify this effect.

Additional work is also needed to refine rubric-sensitive prompting strategies, especially for linguistic micro-level features such as vocabulary, syntax, and punctuation, where agreement gaps remain larger. Exploring hybrid approaches that combine LLM-based reasoning with deterministic NLP tools represents a promising direction.

Overall, the findings provide empirical evidence that carefully designed Human-in-the-Loop AI scoring systems can support large-scale writing assessment processes. With continued validation and responsible deployment, such approaches have the potential to improve scalability, timeliness, and consistency in educational assessment contexts.

\section*{Acknowledgements}
We gratefully acknowledge the work of the expert human raters for their participation.

\section*{Declarations}
\textbf{Funding:} This research received no specific grant from any funding agency in the public, commercial, or not-for-profit sectors.\\
\textbf{Data privacy management:} The study was conducted using secondary assessment data provided by the National Public Education Administration (ANEP) of Uruguay. The dataset made available to the authors was fully anonymized and contained only the written responses together with their corresponding human-assigned scores. No personally identifiable information was included in the data used for this research. LLM-based evaluations were performed through the OpenAI Enterprise API. According to the enterprise service terms, data submitted through the API are not retained for model training or used to improve future OpenAI models, thereby preserving the confidentiality of the assessment data.\\
\textbf{Data availability:} The dataset analyzed in this study consists of responses from an operational nationwide assessment administered by the National Public Education Administration (ANEP) of Uruguay. Due to confidentiality restrictions, the dataset is not publicly available.\\

\bibliography{soa_acredita}

%%\bibliography{sn-bibliography}% common bib file
%% if required, the content of .bbl file can be included here once bbl is generated
%%\input sn-article.bbl

\end{document}